\documentclass[]{spie}  

\usepackage{amsmath,amsfonts,amssymb}
\usepackage{graphicx}
\usepackage[colorlinks=true, allcolors=blue]{hyperref}

\title{Automated external cervical resorption segmentation in cone-beam CT using local texture features}

\author[a]{Sadhana Ravikumar}
\author[b]{Asma A. Khan}
\author[c]{Matthew C. Davis}
\author[d]{Beatriz Paniagua}
\affil[a,d]{Kitware Inc., Carrboro, NC, USA}
\affil[b]{UT Health San Antonio, San Antonio, TX, USA}
\affil[c]{Private Practice in Endodontics, Winnetka, IL, USA, }

\authorinfo{Send correspondence to B.P.\\B.P.: E-mail: beatriz.paniagua@kitware.com}

\begin{document} 
\maketitle

\begin{abstract}
External cervical resorption (ECR) is a resorptive process affecting teeth. While in some patients, active resorption ceases and gets replaced by osseous tissue, in other cases, the resorption progresses and ultimately results in tooth loss. For proper ECR assessment, cone-beam computed tomography (CBCT) is the recommended imaging modality, enabling a 3-D characterization of these lesions. While it is possible to manually identify and measure ECR resorption in CBCT scans, this process can be time intensive and highly subject to human error. Therefore, there is an urgent need to develop an automated method to identify and quantify the severity of ECR resorption using CBCT. Here, we present a method for ECR lesion segmentation that is based on automatic, binary classification of locally extracted voxel-wise texture features. We evaluate our method on 6 longitudinal CBCT datasets and show that certain texture-features can be used to accurately detect subtle CBCT signal changes due to ECR. We also present preliminary analyses clustering texture features within a lesion to stratify the defects and identify patterns indicative of calcification. These methods are important steps in developing prognostic biomarkers to predict whether ECR will continue to progress or cease, ultimately informing treatment decisions.
\end{abstract}


\section{INTRODUCTION}
\label{sec:intro}  

External cervical resorption (ECR) is a resorptive disease affecting teeth that can potentially lead to tooth loss. ECR lesions appear within the dentin, initiating at the periodontal attachment, and its severity can range from small regions on the root surface to more extensive invasion into deeper tooth structures. Although the exact etiology of ECR remains unknown, ECR has been associated with orthodontics, trauma, surgery, periodontal therapy and internal bleaching \cite{goodell2018impact, heithersay2004invasive}. Despite its destructive nature, timely diagnosis of ECR is challenging for clinicians since patients with ECR are commonly asymptomatic. In many cases inflammatory osteoclastic activity slows or ceases which is then often followed by osteoblastic repair of the defect with bone-like tissue. In these cases, the tooth may remain asymptomatic and healthy indefinitely \cite{Mavridou2016}. There is currently no way to predict which ECR lesions will reach this healed stasis and which will display progressive invasion \cite{chen2021review}. Improved prognostication of ECR lesions could assist in deciding which lesions to monitor and which ones require intervention.


Radiographic examinations play a major role in the determination of the extent of ECR and disease prognosis. In recent years, cone-beam computed tomography (CBCT) has become the recommended imaging modality for ECR diagnosis due to is widespread availability and its ability to provide 3-D visualization of the affected tooth \cite{matny2020volumetric,patel2009detection}. CBCT has vastly improved the characterization and localization of ECR. While it is possible to manually identify and measure ECR resorption in CBCT images for disease diagnosis and monitoring, this process is laborious, error prone, subjective and time consuming. Therefore, there is an urgent need to develop an automated method to identify and quantify the severity of ECR resorption in CBCT images \cite{matny2020volumetric}. 

To address this gap, in this paper we present a novel method based on texture features and support vector machines (SVMs) to automatically segment ECR lesions using high-resolution CBCT data. Since textural biomarkers are based on the statistical properties of the image intensities and not directly on the intensity values themselves, we believe the proposed approach will be robust to CBCT images collected from different settings. We evaluate our approach on longitudinal data acquired from three patients and present preliminary findings suggesting that texture features may be helpful in both isolating the ECR lesion and distinguishing radiolucent, osteolytic areas from radiopaque, osteogenic regions within ECR lesions.

\section{Materials}
\label{sec:title}
Our study uses longitudinal, high-resolution dental CBCT scans acquired at two different time-points between 2014 and 2024 from 3 patients with an ECR diagnosis, resulting in a total of 6 scans. The average time between the two time-points is 4.67 $\pm$ 3.86 years. The dental imaging system uses a Carestream CBCT Unit (90 kVp, 10 mA, 50 mm x 50 mm field of view), with an isotropic voxel size of 76 $\mu$m. 

\section{METHODS}
\begin{figure} [ht]
   \begin{center}
   \begin{tabular}{c} 
   \includegraphics[height=5cm]{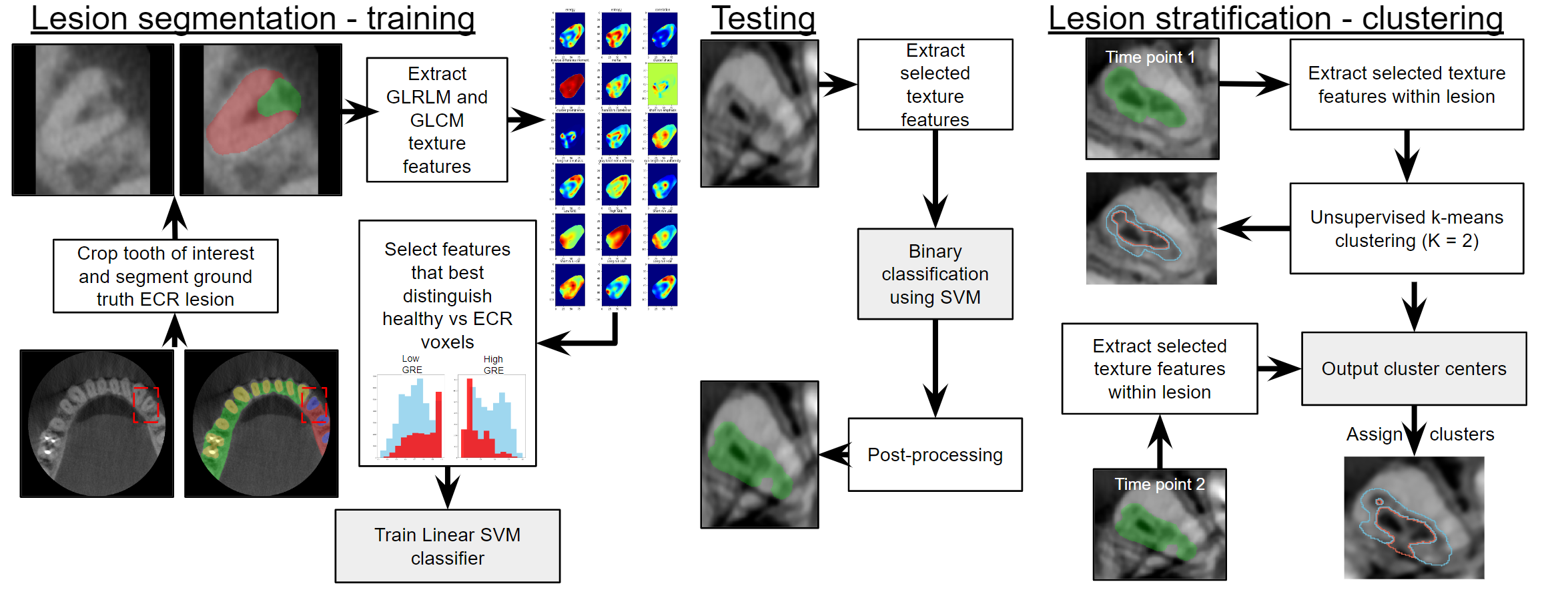}
   \end{tabular}
   \end{center}
   \caption[example] 
   { \label{fig:schematic} 
Schematic overview of the proposed ECR lesion segmentation and stratification approach.}
   \end{figure}

The presented algorithm for ECR lesion segmentation is based on automatic, binary classification of locally extracted voxel-wise texture features and lesion stratification is achieved using unsupervised k-means clustering. The following sections describe the methods in more detail. Figure \ref{fig:schematic} provides an schematic overview of the proposed approach. 

 
\subsection{Dental segmentation}
 In each scan, we performed dental segmentation using the ``DentalSegmentator" extension available in 3D Slicer \cite{Kikinis2014,dot2024dentalsegmentator,isensee2021nnu}. We extracted cropped volumes containing the ECR-affected tooth from both the CBCT scan and corresponding dental segmentation. The cropped tooth segmentations were manually edited to correct any errors. An expert rater then manually generated ground truth ECR lesion segmentations. 
 

\subsection{Local texture feature extraction}
Prior to feature extraction, the cropped CBCT scans were pre-processed by clipping the intensity range to lie within the 5$^{th}$ and 95$^{th}$ percentiles to exclude outlier intensity values due to image artifacts. Scans were then normalized to range [0,1] and Gaussian smoothing (kernel size = 1 voxel) was applied to minimize noise.

Our group pioneered an implementation of traditional texture analysis methods \cite{Haralick1973,Galloway1975} where it is possible to compute local texture feature maps that describe the statistical properties of image intensities at each voxel in relation to its neighboring voxels. This work was packaged in the Insight Toolkit (ITK) module itkTextureFeatures\cite{vimort2017computing}, that we used to extract texture-based features within the segmented tooth region. This module computes eight features based on the gray-level co-occurrence matrix (GLCM) and ten features based on the grey-level run length matrix (GLRLM). The GLCM matrix describes intensity organization of each voxels' neighborhood and the GLRLM matrix is based on the gray-level run which is a set of consecutive, collinear voxels having the same gray-level value. For both filters, we conducted experiments using different values for the neighborhood radius parameter (2, 5, 7 and 9 voxels) which determines the scale of the textures targeted. Since the texture features are extracted for each voxel, it results in a large amount of training samples, most of which belong to the healthy tooth class, which can lead to over-fitting. Therefore, we computed the smallest rectangular region that contains all of the lesion voxels in the segmentation and used it to crop the feature maps, leaving a 5-voxel margin of background on all sides. For each texture feature, we visually examined histograms comparing the distribution of feature values for healthy and ECR lesion voxels and found that the low grey level run emphasis (LGRE) and high grey level run emphasis (HGRE) feature maps provide the greatest separation between healthy and pathological tissue. Therefore, LGRE and HGRE features were selected for classifier training. 


\subsection{Support vector machine segmentation}

We implemented a Linear SVM classifier using the Python scikit-learn library. Before training, we standardized the training features by removing the mean and scaling to unit variance, and then applied the same scaling parameters to the feature vectors extracted during the testing phase. The segmentation output by the SVM classifier tends to over-segment the lesion and may still contain mis-clasified voxels. To counter-balance the over-segmentation, we first applied binary erosion (kernel size = 6 voxels) before extracting the largest connected component. We then applied binary dilation, with the same kernel size, to obtain the final output segmentation. 

\section{RESULTS}
\label{sec:sections}
\begin{figure} [ht]
   \begin{center}
   \begin{tabular}{c} 
   \includegraphics[height=5cm]{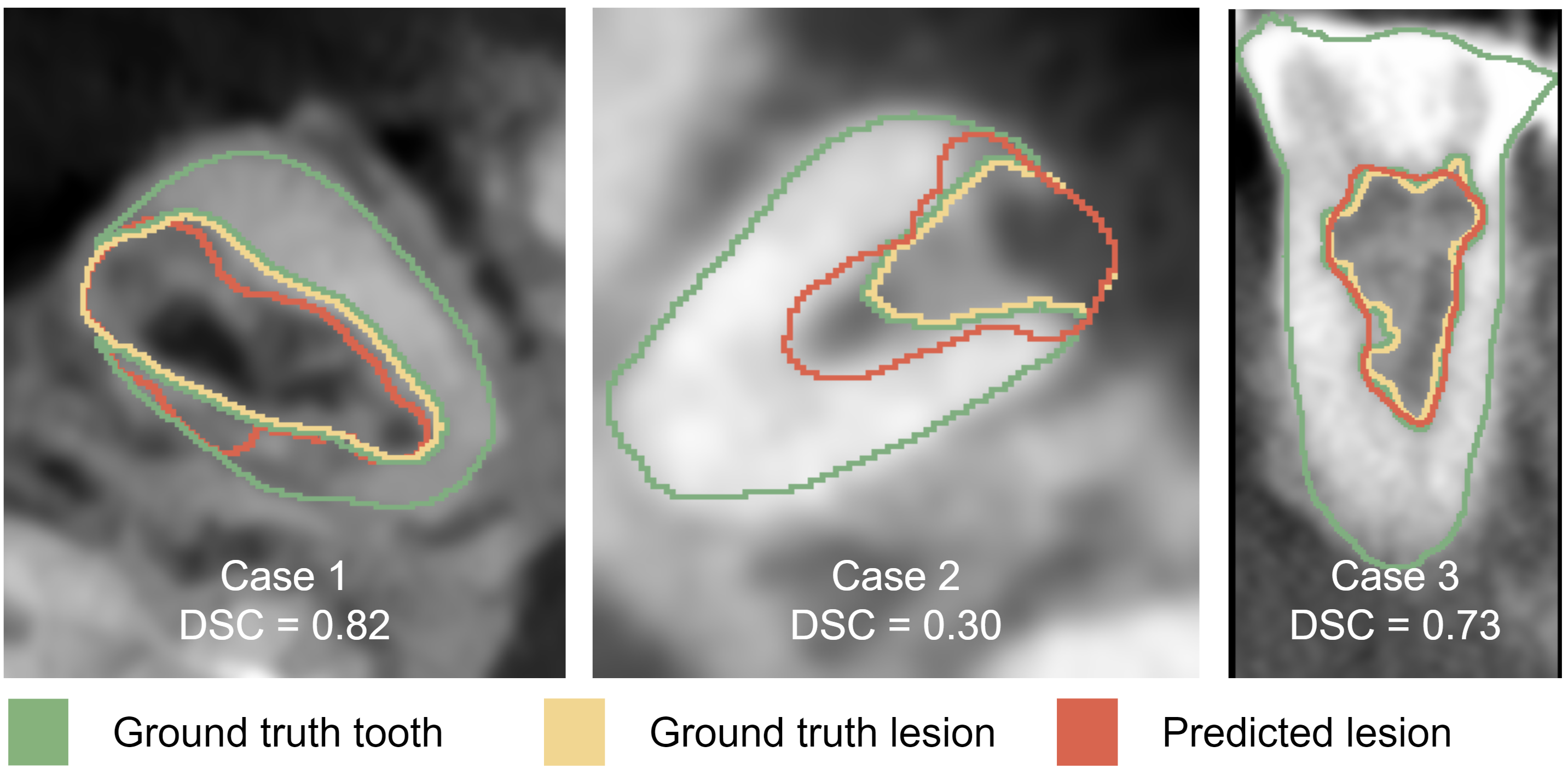}
   \end{tabular}
   \end{center}
   \caption[example] 
   { \label{fig:result} 
Cross-sectional views of example predictions obtained for each patient overlaid on the corresponding CBCT scan. The ground truth ECR lesion is outlined in yellow, and the predicted segmentation is shown in red. The example scans shown for Case 1 and Case 2 achieved the best and worst Dice score coefficients (DSC) respectively.}
   \end{figure}

We evaluated our method by performing leave-one-out cross validation at the patient level. In each split, we trained the SVM using data from two patients and evaluated on data from the left out patient. We report precision and recall, weighted by the class support and averaged over all six cases. We also compute the Dice Score Coefficient (DSC) between the ground truth and predicted ECR segmentation following post-processing (Table \ref{tab:fonts}). Based on these results, we selected the optimal neighborhood radius for feature extraction as 5 voxels. Figure \ref{fig:result} shows example cross-sectional views from each patient, including the best and worst performing cases based on DSC. Three scans achieved a DSC $>$ 0.7. Qualitatively, we see the proposed method is able to predict reasonably accurate lesion segmentations. Although scans from case 2 resulted in poor DSC metrics, thus lowering the average metric, we observe that this is due to the over-segmentation including the tooth pulp which is similar in appearance to ECR lesions (Figure \ref{fig:result} middle). 

\begin{table}[ht]
\caption{Average dice score coefficient, precision and recall achieved on the held out test set for experiments conducted with different texture feature neighborhoods.} 
\label{tab:fonts}
\begin{center}       
\begin{tabular}{|c|c|c|c|} 
\hline
\rule[-1ex]{0pt}{3.5ex}  \textbf{Neighborhood Radius} & \textbf{Dice Score Coefficient} & \textbf{Precision} & \textbf{Recall}\\
\hline
\rule[-1ex]{0pt}{3.5ex}  2 & 0.56 $\pm$ 0.18 &  0.86 $\pm$ 0.04 & 
 0.83$\pm$ 0.05  \\
\hline
\rule[-1ex]{0pt}{3.5ex} 5 &   0.59 $\pm$ 0.19 & 0.86 $\pm$ 0.04 & 0.84 $\pm$ 0.05\\
\hline
\rule[-1ex]{0pt}{3.5ex}  7 &  0.59 $\pm$ 0.19 & 0.86 $\pm$ 0.04 & 0.84 $\pm$ 0.05 \\
\hline
\rule[-1ex]{0pt}{3.5ex}  9 & 0.47 $\pm$ 0.33  & 0.85 $\pm$ 0.05 & 0.83 $\pm$ 0.06   \\
\hline
\end{tabular}
\end{center}
\end{table} 

\begin{figure} [ht]
   \begin{center}
   \begin{tabular}{c} 
   \includegraphics[height=5cm]{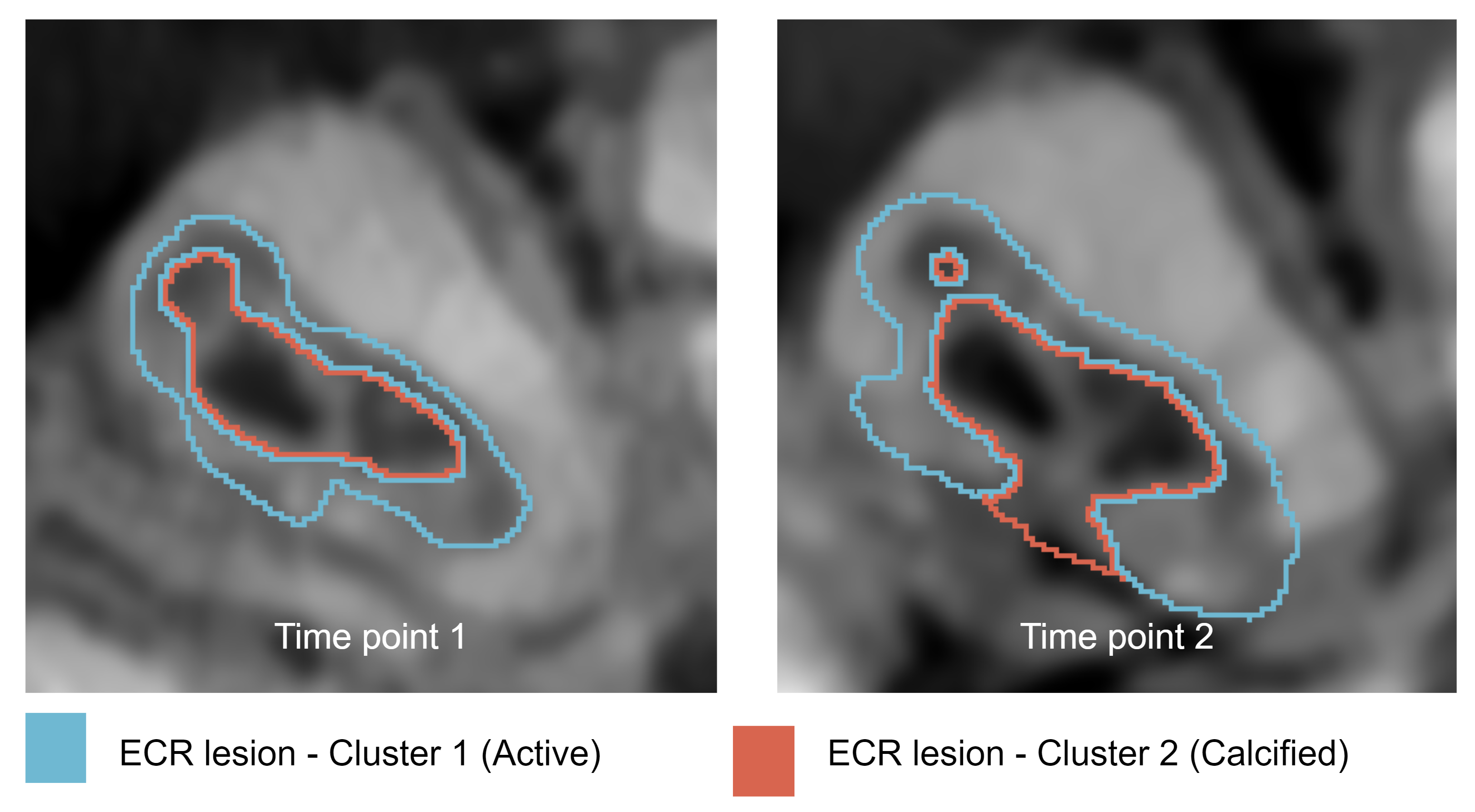}
   \end{tabular}
   \end{center}
   \caption[example] 
   { \label{fig:clusters} 
Automated ECR lesion segmentations for Case 1 predicted at two time-points 1 year apart. Unsupervised k-means clustering identifies a distinct, darker inner cluster within the lesion that may correspond to calcified tissue.}
   \end{figure}

We performed an exploratory analysis using predictions obtained for Case 1 looking at within lesion texture patterns, computed across the two time-points as shown in Figure \ref{fig:clusters}. We performed unsupervised k-means clustering (k = 2) using LGRE and HGRE texture features extracted from lesion voxels at the first time-point and used the computed cluster centers to predict cluster assignments on features extracted from the second time-point. At both timepoints, we observe a darker, radiopaque cluster within the lesion which likely corresponds to calcified tissue. Overall, the second scan shows increased calcified tissues but still presents with larger radiolucency areas (ECR lesion volume=134.74 mm$^3$) compared with the first scan (ECR lesion volume=77.98 mm$^3$).

\section{CONCLUSION}
In this paper, we introduce a novel method for automated ECR lesion segmentation and stratification based on local texture features. This is an important step in being able to better monitor ECR progression and predict clinical outcomes. Our results indicate that the computed texture features are able to identify subtle differences in CBCT signal that reflect differences in dentin integrity, and can be used to characterize ECR lesion variability. The main limitation of the current study is the small sample size. Future work includes developing an SVM lesion segmentation model using a larger CBCT dataset. With a larger dataset, we will also be able to perform analyses associating within lesion texture features with clinical outcomes with the goal of identifying imaging biomarkers for improved ECR prognosis that will help inform treatment decisions. 
 


\bibliography{report} 
\bibliographystyle{spiebib} 

\end{document}